\documentclass{article}

\usepackage{graphicx}
\usepackage{algorithm}
\usepackage{amsmath}
\usepackage{algpseudocode}

 \usepackage[preprint]{neurips_2025}

\usepackage[utf8]{inputenc} 
\usepackage[T1]{fontenc}    
\usepackage{hyperref}       
\usepackage{url}            
\usepackage{booktabs}       
\usepackage{amsfonts}       
\usepackage{nicefrac}       
\usepackage{microtype}      
\usepackage{xcolor}         

\title{Frame-based Equivariant Diffusion Models for 3D Molecular Generation}

\author{%
    Mohan Guo\thanks{Equal contribution.} \\
    {University of Amsterdam} \\
    \texttt{mohan.guo@student.uva.nl} \\
    \And
    Cong Liu\footnotemark[1] \\
    {AMLab} \\
    {University of Amsterdam} \\
    \texttt{c.liu4@uva.nl} \\
    \AND
    Patrick Forré \\
    {AI4Science Lab} \\
    {Korteweg-de Vries Institute for Mathematics} \\
    {University of Amsterdam} \\
  \texttt{p.d.forre@uva.nl} \\
}

\begin{document}

\maketitle

\begin{abstract}
Recent methods for molecular generation face a trade-off: they either enforce strict equivariance with costly architectures or relax it to gain scalability and flexibility. 
We propose a frame-based diffusion paradigm that achieves deterministic $\mathbb{E}(3)$-equivariance while decoupling symmetry handling from the backbone. 
Building on this paradigm, we investigate three variants: Global Frame Diffusion (GFD), which assigns a shared molecular frame; Local Frame Diffusion (LFD), which constructs node-specific frames and benefits from additional alignment constraints; and Invariant Frame Diffusion (IFD), which relies on pre-canonicalized invariant representations.
To enhance expressivity, we further utilize EdgeDiT, a Diffusion Transformer with edge-aware attention. 

On the QM9 dataset, GFD with EdgeDiT achieves state-of-the-art performance, with a test NLL of $-137.97 \pm 0.00$ at standard scale and $-141.85 \pm 0.00$ at double scale, alongside atom stability of $98.98\%$, and molecular stability of $90.51\%$. 
These results surpass all equivariant baselines while maintaining high validity and uniqueness and nearly $2\times$ faster sampling compared to EDM. 
Altogether, our study establishes frame-based diffusion as a scalable, flexible, and physically grounded paradigm for molecular generation, highlighting the critical role of global structure preservation.

\end{abstract}

\section{Introduction}

Molecular design is central to drug discovery and materials science, but traditional trial-and-error approaches are costly and slow \citep{dimasi2016innovation}, motivating the need for computational methods.
Recent advances in deep learning have enabled accurate molecular property prediction \citep{luo2021predicting, zhou2023uni, fang2022geometry}, but such models remain limited to evaluating existing compounds. Generative models aim to overcome this limitation by designing novel, chemically valid molecules. Among these, diffusion models have shown particular effectiveness in modeling complex distributions. By learning to reverse a gradual noise corruption process, they enable high-quality sample generation \citep{song2019generative, ho2020denoising, hoogeboom2022equivariant}.
Diffusion models have achieved remarkable success in domains such as images \citep{ho2020denoising, rombach2022high} and videos \citep{ho2022video, blattmann2023stable}, demonstrating their ability to model high-dimensional, continuous, and structured data. These characteristics align closely with molecular generation, where data consists of 3D atomic coordinates and intricate structural patterns.

A central challenge is to incorporate symmetries into generative models. Molecules are structured as atoms in continuous 3D space connected by bonds, requiring joint modeling of geometry and relational structure. Graph neural networks (GNNs) are well suited for molecular graphs, achieving permutation equivariance \citep{scarselli2008graph}. Extensions such as Equivariant Graph Neural Networks (EGNNs) use pairwise distances as invariant features and update coordinates equivariantly, making them efficient and widely adopted in diffusion-based molecular models \citep{satorras2021n}. Multi-Channel EGNN (MC EGNN) extends EGNNs by maintaining multiple coordinate channels throughout message passing, parameterizing the number of input, hidden, and output channels to enrich representation power \citep{levy-icml2023}.
Equivariant Diffusion Model (EDM) employs EGNNs as the backbone models in the diffusion model to respect 3D symmetries \citep{hoogeboom2022equivariant}. GeoLDM further extends this by operating in latent space \citep{xu2023geometric}. 
Despite their effectiveness, GNN-based frameworks couple equivariance tightly with message passing, limiting architectural flexibility. They inherit known issues, including restricted expressivity due to the 1-Weisfeiler–Lehman bound \citep{balcilar2021breaking}, and over-squashing of long-range dependencies \citep{alonbottleneck}.
These factors hinder scalability and efficiency in large systems. 

Canonicalization methods achieve equivariance by mapping inputs into canonical representations, often through learned local frames \citep{Luo_2022_CVPR, kaba2023equivariance}. They offer universal approximation guarantees and can be integrated with diverse backbones via lightweight modules \citep{ma2024a}, with recent refinements in local frame construction \citep{lippmann2025beyond}. However, their application to generative models such as diffusion remains limited.

Canonicalization-based model Canon addresses symmetries by aligning data to a canonical orientation prior to diffusion \citep{DBLP:journals/corr/abs-2501-07773}. In our study, we also explored a related idea through the IFD framework. However, both canon and IFD reduce the intrinsic diversity of the diffusion noise process, which limits their generative performance compared to approaches that preserve this diversity, such as GFD and LFD.

Rotationally Aligned Diffusion Model (RADM) elaxes strict equivariance constraints by learning a rotationally aligned latent space. This design allows diffusion to be performed with non-equivariant backbones such as standard GNNs or Transformers, thereby improving scalability and efficiency compared to equivariant diffusion models. However, RADM still introduces additional complexity through its alignment stage and does not achieve the same level of generative performance as our proposed GFD framework \citep{dingscalable}.

SymDiff offers a more flexible alternative via stochastic symmetrization \citep{zhang2025symdiff}. It enforces $\mathbb{E}(3)$-equivariance through random symmetry transformations during the reverse diffusion process, allowing scalable backbones like Diffusion Transformers. Its reliance on stochastic sampling, however, risks biased symmetry coverage and reduced robustness to unseen orientations.


In this work, we introduce a frame-based diffusion paradigm that achieves deterministic $\mathbb{E}(3)$-equivariance while enabling flexible backbone architectures. 
We investigate three architectural variants: Global Frame Diffusion (GFD), which constructs a shared molecular frame; Local Frame Diffusion (LFD), which assigns atom-specific frames; and Invariant Frame Diffusion (IFD), which directly operates on invariant representations. 
Our study shows that while unconstrained LFD underperforms due to disrupted global consistency, introducing a frame alignment constraint markedly improves its performance, thereby validating our hypothesis that preserving global Euclidean structure is essential. 
For the backbone, we adopt the Diffusion Transformer (DiT) \citep{peebles2023scalable}. By incorporating edge information into the attention mechanism, the resulting EdgeDiT achieves improved performance.
Among these variants, GFD combined with EdgeDiT achieves the strongest empirical results on QM9, with a test NLL of $-137.97 \pm 0.00$, atom stability of 98.89\%, and molecular stability of 89.39\% at standard scale suggested in SymDiff. 
At double scale, GFD further improves to an NLL of $-141.85 \pm 0.00$, atom stability of 98.98\%, and molecular stability of 90.51\%, consistently surpassing prior equivariant baselines while reducing sampling time by nearly half compared with EDM.
Together, these findings establish frame-based diffusion as a scalable, expressive, and physically grounded paradigm for molecular generation, bridging theoretical guarantees with practical efficiency.

\section{Methodology}
We achieve deterministic $\mathbb{E}(3)$-equivariance in diffusion-based molecular generation by projecting atomic environments into frames, ensuring that the backbone processes only invariant representations.

Each molecule consists of atomic coordinates $\mathbf{x}=(\mathbf{x}_1,...,\mathbf{x}_N)$ and features $\mathbf{h}=(\mathbf{h}_1,...,\mathbf{h}_N)$ of its $N$ atoms, with $\mathbf{x}_i \in \mathbb{R}^3$ and $\mathbf{h}_i \in \mathbb{R}^{d_h}$. 
We represent the combination of coordinates and features of every node as $\mathbf{m}=(\mathbf{m}_1,\dots,\mathbf{m}_N)$, where $\mathbf{m}_i=[\mathbf{x}_i; \mathbf{h}_i]$ for atom $i$.
Since $\mathbf{h}$ is invariant to translation and rotation, only coordinate equivariance needs to be addressed.

A transformation $g \in \mathbb{E}(3)$ is defined by $(\mathbf{R}, \mathbf{t})$ with $\mathbf{R} \in O(3)$ and $\mathbf{t} \in \mathbb{R}^3$.
Under $g$, atomic coordinates transform as
$(\mathbf{x}, \mathbf{h}) \to (\mathbf{Rx} + \mathbf{t}, \mathbf{h})$.
In group theory, a group $\mathbb{G}$ acts on a space $\mathcal{X}$ through group actions $\mathcal{T}_g : \mathcal{X} \to \mathcal{X}$ for each $g \in \mathbb{G}$. 
A function $f: \mathcal{X} \to \mathcal{Y}$ is $\mathbb{G}$-equivariant if there exist corresponding group actions $\mathcal{S}_g : \mathcal{Y} \to \mathcal{Y}$ satisfying:
\begin{equation}
f(\mathcal{T}_g(x)) = \mathcal{S}_g(f(x))
\quad
\forall g \in \mathbb{G},  \forall x \in \mathcal{X} 
\end{equation}

In this work, we considered equivariance over $\mathbb{E}(3)$ and permutation group.
For translation equivariance, we follow EDM by centering molecules at the zero-mass position, preserved under zero-centered noise, ensuring translation invariance \citep{hoogeboom2022equivariant, xu2022geodiff}. 
For permutation equivariance, we remove positional encodings in DiTs, and since MC-EGNN is already permutation-equivariant, this property is naturally satisfied.


\begin{algorithm}
\caption{Frame-based Equivariant Projection (single diffusion step)}
\label{alg:frame_projection}
\begin{algorithmic}[1]
\Require Noisy atomic coordinates $\mathbf{x}=\{\mathbf{x}_i\}_{i=1}^N$, features $\mathbf{h}=\{\mathbf{h}_i\}_{i=1}^N$, equivariant frame constructor $\phi_e$, backbone $\phi_\theta$
\Ensure Equivariant outputs $\mathbf{y}^{\text{out}}=\{\mathbf{y}_i^{\text{out}}\}_{i=1}^N$
  \State Construct orthogonal frame $ \phi_e(\{\mathbf{x}_i,\mathbf{h}_i\}_{i=1}^N)  \rightarrow\{\mathbf{O}_i\}_{i=1}^N = \{[\mathbf{u}_{i1},\mathbf{u}_{i2},\mathbf{u}_{i3}]\}_{i=1}^N $ 
\For{each atom $i=1,\dots,N$}
  \State \textit{Equivariance: under rotation $\mathbf{R}$, we have $\mathbf{O}_i'=\mathbf{R}\mathbf{O}_i$}
  \State Project coordinates into frame: $\mathbf{x}_i^{\text{in}} = \mathbf{O}_i^{-1}\mathbf{x}_i$
\EndFor
\State Backbone prediction on invariant inputs: $\mathbf{y} = \phi_\theta(\{\mathbf{x}_i^{\text{in}}, \mathbf{h}_i\}_{i=1}^N)$
\For{each atom $i=1,\dots,N$}
  \State Map back to original coordinates: $\mathbf{y}_i^{\text{out}} = \mathbf{O}_i\, \mathbf{y}_i$
\EndFor
\State \Return $\mathbf{y}^{\text{out}}$
\end{algorithmic}
\end{algorithm}

\subsection{Local Frame-based Diffusion Model}
\begin{figure}[htbp]
    \centering
    \includegraphics[width=0.7\linewidth]{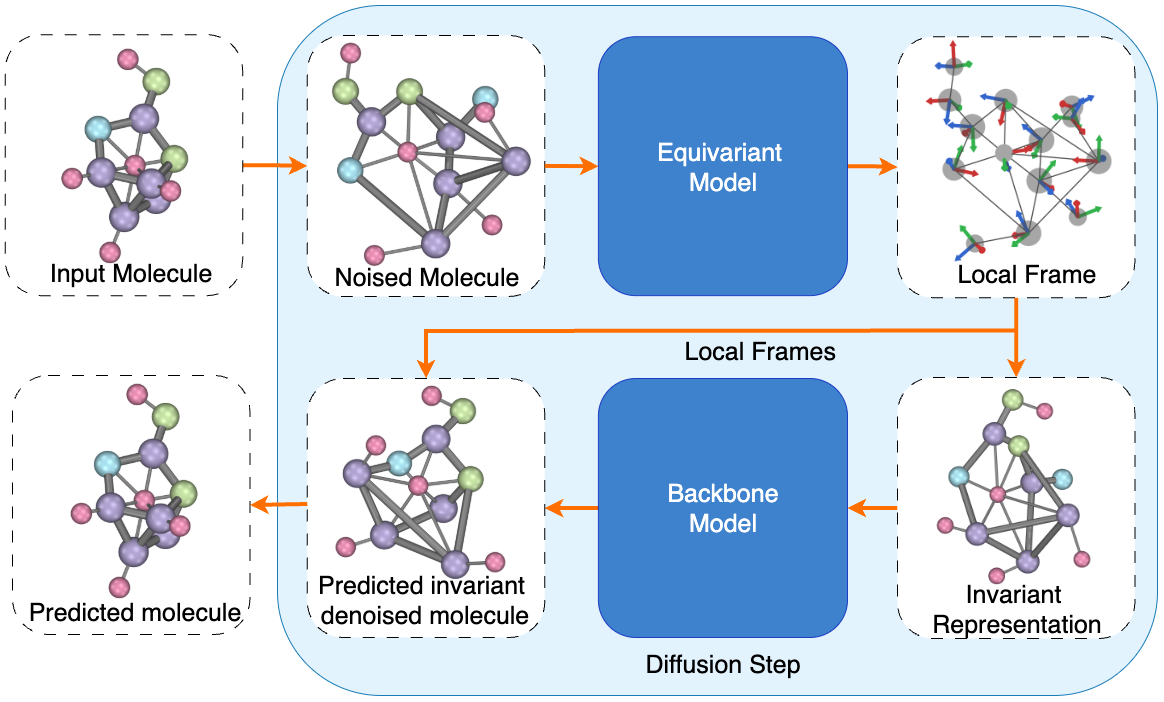}
    \caption{Local Frame-based Diffusion Model (LFD). The framework operates through the following process: (1) The input molecule is noised and then is processed by an equivariant model to construct local frames for each atom. (2) These local frames are used to derive invariant representations that capture local molecular geometry while being invariant to $\mathbb{E}(3)$ group. (3) The backbone diffusion model takes the invariant representations as inputs to predict the invariant denoised molecule. (4) Predicted molecule is obtained by applying the local frames inversely. Loss is computed between input molecule and predicted molecule.}
    \label{fig:frame-local}
\end{figure}

As illustrated in Figure~\ref{fig:frame-local}, LFD relies on an equivariant module to construct local frames for each atom. By projecting noisy coordinates into local coordinate systems, the model obtains invariant representations with respect to $\mathbb{E}(3)$ and permutation transformations. These invariant features are processed by the backbone diffusion model, and predictions are mapped back into global space via the local frames, enabling equivariance.

The training procedure is summarized in Algorithm~\ref{alg:lfd-train}, where the backbone learns to denoise projected coordinates. Sampling follows the reverse process (Algorithm~\ref{alg:lfd-sample}). 

\begin{algorithm}
\caption{Local Frame-based Diffusion (LFD) - Training}
\label{alg:lfd-train}
\begin{algorithmic}[1]
\Require Molecule $\mathbf{m}$
\State Sample $t \sim \text{Uniform}(1, T)$, $\boldsymbol{\epsilon} \sim \mathcal{N}(\mathbf{0}, \mathbf{I})$
\State $\mathbf{z}_t = \alpha_t \mathbf{m} + \sigma_t \boldsymbol{\epsilon}$
\State $\{\mathbf{O}_i\}_{i=1}^N = \phi_e(\mathbf{z}_t)$ \Comment{Equivariant frames}
\State Project: $\mathbf{z}_t^{\text{local}} = \{\mathbf{O}_i^{-1}\mathbf{z}_{t,i}\}$
\State Predict: $\hat{\boldsymbol{\epsilon}}^{\text{local}} = \phi_\theta(\mathbf{z}_t^{\text{local}}, t)$
\State Invert: $\hat{\boldsymbol{\epsilon}}_i = \mathbf{O}_i \hat{\boldsymbol{\epsilon}}^{\text{local}}_i$
\State Minimize $\mathcal{L} = \|\boldsymbol{\epsilon} - \hat{\boldsymbol{\epsilon}}\|^2$
\end{algorithmic}
\end{algorithm}

\begin{algorithm}
\caption{Local Frame-based Diffusion (LFD) - Sampling}
\label{alg:lfd-sample}
\begin{algorithmic}[1]
\State Sample $\mathbf{z}_T \sim \mathcal{N}(\mathbf{0}, \mathbf{I})$
\For{$t = T, T-1, ..., 1$}
    \State $\{\mathbf{O}_i\} = \phi_e(\mathbf{z}_t)$
    \State Project $\mathbf{z}_t$ into local frames, predict noise, invert
    \State Update $\mathbf{z}_{t-1}$ via reverse diffusion step
\EndFor
\State \Return $\mathbf{z}_0$
\end{algorithmic}
\end{algorithm}


One central hypothesis of this work is that the superior performance of GFD compared to LFD arises from its preservation of inter-atomic relations in the Euclidean space, which maintains the natural mapping between molecular geometry and physico-chemical properties. To examine this hypothesis, we extend LFD with an additional frame alignment constraint that encourages local frames to remain consistent with a global molecular frame. 
Concretely, given rotation matrices $\mathbf{O}_i, \mathbf{O}_g \in SO(3)$, their relative rotation is
\[
\mathbf{R}_{i,g} = \mathbf{O}_i^\top \mathbf{O}_g.
\]
The angle $\theta_{i,g}$ corresponding to $\mathbf{R}_{i,g}$ can be obtained from
\[
\cos \theta_{i,g} = \tfrac{1}{2} \big( \mathrm{tr}(\mathbf{R}_{i,g}) - 1 \big), 
\quad
\sin \theta_{i,g} = \tfrac{1}{2} \lVert \mathbf{R}_{i,g} - \mathbf{R}_{i,g}^\top \rVert_F.
\]
We then compute
\[
\theta_{i,g} = \arctan2(\sin \theta_{i,g}, \cos \theta_{i,g}),
\]
which gives the geodesic distance on $SO(3)$ between local and global frames. The alignment loss is defined as
\[
\mathcal{L}_{\text{align}} = \frac{1}{N} \sum_{i=1}^N \frac{\theta_{i,g}}{\pi}.
\]

This term is added to the diffusion training loss to encourage consistency between local and global orientations with a weight:
\[
\mathcal{L} = \mathcal{L}_{\mathrm{diff}} + \lambda \, \mathcal{L}_{\mathrm{align}},
\]
where $\mathcal{L}_{\mathrm{diff}}$ is the standard diffusion loss and $\lambda$ is a balancing weight.

\subsection{Global Frame-based Diffusion Model}
Another architecture is the Global Frame-based Diffusion Model (GFD), which constructs a single molecular frame to project coordinates into invariant representations before diffusion modeling and restores them via inverse projection. This guarantees exact $\mathbb{E}(3)$-equivariance while preserving global geometric relationships, enabling the backbone to learn consistent geometric–chemical patterns.

\begin{figure}[htbp]
    \centering
    \includegraphics[width=0.7\linewidth]{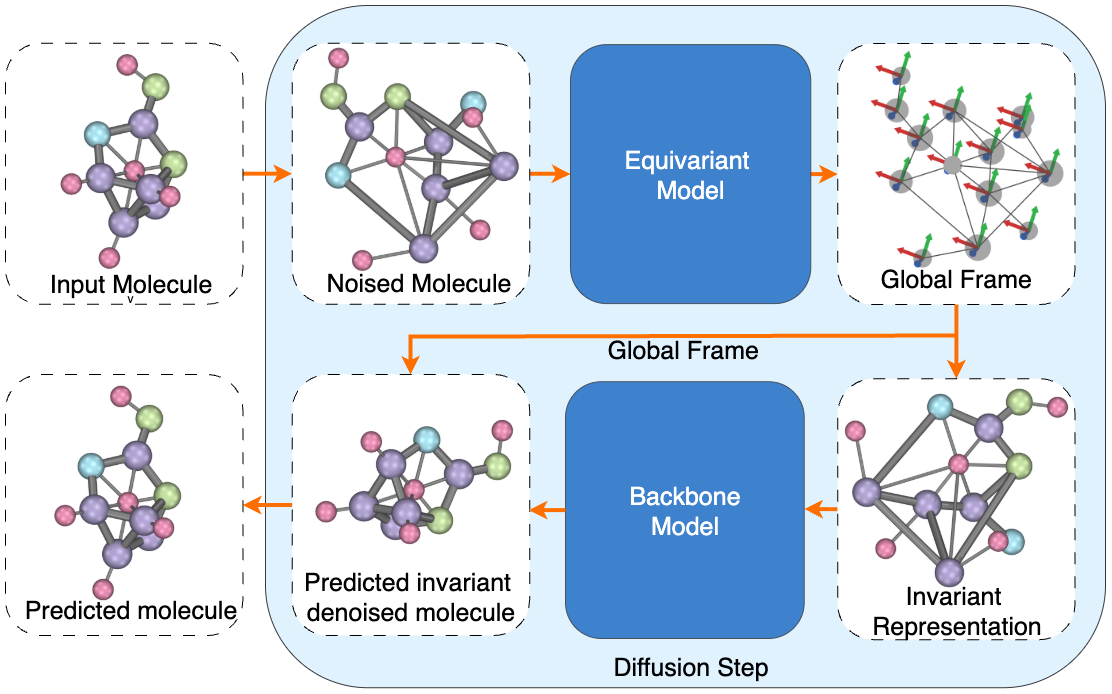}
    \caption{Global Frame-based Diffusion (GFD): an equivariant module constructs a global molecular frame from noised inputs, invariant features are derived and denoised by the backbone, and the final molecule is reconstructed via inverse frame transformation with loss to the original.}
    \label{fig:frame-global}
\end{figure}

\begin{algorithm}
\caption{Global Frame-based Diffusion (GFD) - Training}
\begin{algorithmic}[1]
\Require Molecule $\mathbf{m}$
\Ensure Loss $\mathcal{L}$

\State Sample $t \sim \text{Uniform}(1, T)$, $\boldsymbol{\epsilon} \sim \mathcal{N}(\mathbf{0}, \mathbf{I})$
\State $\mathbf{z}_t = \alpha_t \mathbf{m} + \sigma_t \boldsymbol{\epsilon}$
\State $\{\mathbf{X}_{fi}\}_{i=1}^N = \phi_e(\mathbf{z}_t)$
\State $\mathbf{O}_{\text{global}} = \text{Pool}(\{\text{GramSchmidt}(\mathbf{X}_{fi})\}_{i=1}^N)$
\For{each atom $i$}
   \State $\mathbf{z}_{t,i}^{\text{global}} = \mathbf{O}_{\text{global}}^{-1} \mathbf{z}_{t,i}$
\EndFor
\State $\hat{\boldsymbol{\epsilon}}^{\text{global}} = \phi_\theta(\mathbf{z}_t^{\text{global}}, t)$
\For{each atom $i$}
   \State $\hat{\boldsymbol{\epsilon}}_i = \mathbf{O}_{\text{global}} \hat{\boldsymbol{\epsilon}}_i^{\text{global}}$
\EndFor
\State Minimize $\mathcal{L} = \|\boldsymbol{\epsilon} - \hat{\boldsymbol{\epsilon}}\|^2$
\end{algorithmic}
\end{algorithm}

\begin{algorithm}
\caption{Global Frame-based Diffusion (GFD) - Sampling}
\begin{algorithmic}[1]
\Require None
\Ensure Generated molecule $\mathbf{m}$

\State Sample $\mathbf{z}_T \sim \mathcal{N}(\mathbf{0}, \mathbf{I})$
\For{$t = T, T-1, ..., 1$}
    \State $\{\mathbf{X}_{fi}\}_{i=1}^N = \phi_e(\mathbf{z}_t)$
    \State $\mathbf{O}_{\text{global}} = \text{Pool}(\{\text{GramSchmidt}(\mathbf{X}_{fi})\}_{i=1}^N)$
    \For{each atom $i$}
        \State $\mathbf{z}_{t,i}^{\text{global}} = \mathbf{O}_{\text{global}}^{-1} \mathbf{z}_{t,i}$
    \EndFor
    \State $\hat{\boldsymbol{\epsilon}}^{\text{global}} = \phi_\theta(\mathbf{z}_t^{\text{global}}, t)$
    \For{each atom $i$}
        \State $\hat{\boldsymbol{\epsilon}}_i = \mathbf{O}_{\text{global}} \hat{\boldsymbol{\epsilon}}_i^{\text{global}}$
    \EndFor
    \State Sample $\boldsymbol{\epsilon} \sim \mathcal{N}(\mathbf{0}, \mathbf{I})$ if $t > 1$, else $\boldsymbol{\epsilon} = \mathbf{0}$
    \State $\mathbf{z}_{t-1} = \frac{1}{\alpha_{t|t-1}} \mathbf{z}_t - \frac{\sigma_{t|t-1}^2}{\alpha_{t|t-1} \sigma_t} \hat{\boldsymbol{\epsilon}} + \sigma_{t \rightarrow t-1} \boldsymbol{\epsilon}$
\EndFor
\State \Return $\mathbf{z}_0$
\end{algorithmic}
\end{algorithm}

\subsection{Invariant Frame-based Diffusion Model}
IFD takes a different approach by obtaining the invariant representation before the noising and denoising process.
This design eliminates the need for coordinate transformation during each sampling step, significantly reducing computational overhead.

\begin{figure}[htbp]
    \centering
    \includegraphics[width=0.7\linewidth]{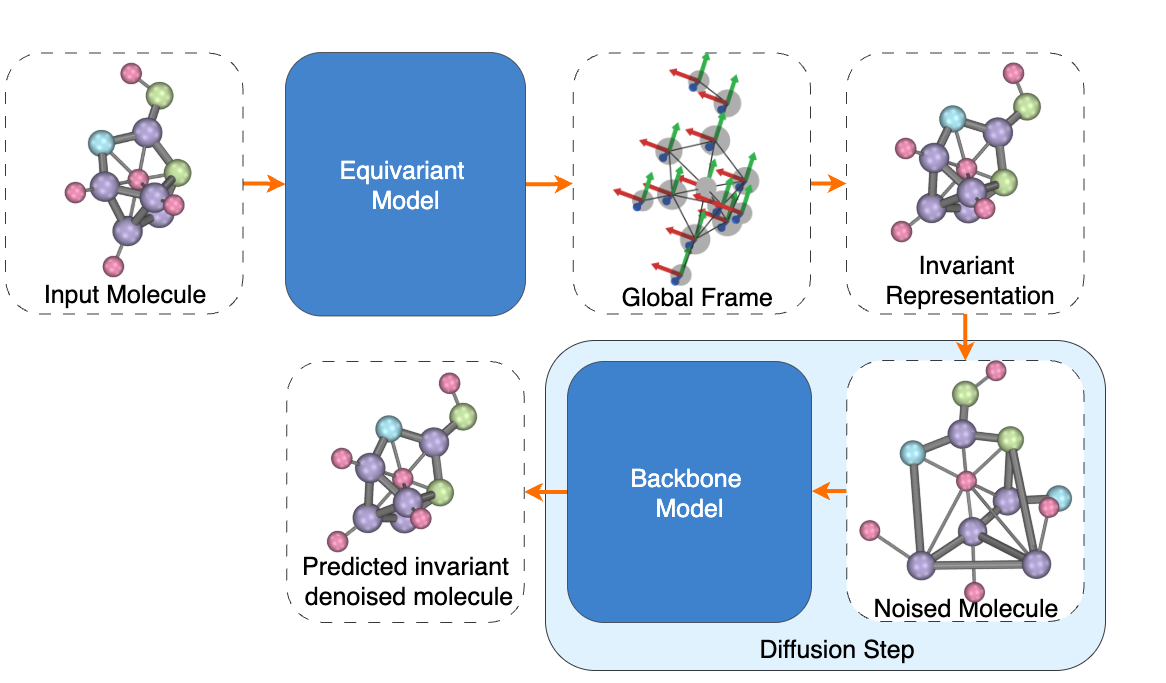}
    \caption{Invariant Frame-based Diffusion Model (IFD). The framework operates in two stages: (1) Pre-canonicalization: An equivariant model processes the input molecule to construct a global frame and derives rotation-invariant representations. (2) Diffusion process: The backbone model takes the noised invariant representation as inputs to predict the denoised molecule structure. The predicted output is the invariant denoised molecule, which ideally should be the rotated input molecule. Loss is computed between invariant representation and predicted invariant denoised molecule..}
    \label{fig:frame-invariant}
\end{figure}

The architecture of IFD is shown in Figure~\ref{fig:frame-invariant}.
The input molecule is passed directly into the equivariant model to obtain the frames of every node. 
Then, a global frame is computed by pooling. 
All atomic coordinates are projected into this global frame to obtain the invariant representation. 
This canonization process is the same as in GFD, except that the invariant representation here is a rotated version of the input molecule rather than the noised molecule. 

The invariant representation is passed into the forward noising process and the backward denoising process of the diffusion model. 
The backbone model predicts the denoised invariant molecule from the input noised invariant representation.
Since we already have the invariant representation, which is also a valid molecule, the training loss can be computed directly between the prediction and the invariant representation. 

\begin{algorithm}
\caption{Invariant Frame-based Diffusion (IFD) - Training}
\begin{algorithmic}[1]
\Require Clean molecule $\mathbf{m}$
\Ensure Loss $\mathcal{L}$

\State \textbf{Pre-canonization:}
\State $\{\mathbf{X}_{fi}\}_{i=1}^N = \phi_e(\mathbf{m})$
\State $\mathbf{O}_{\text{global}} = \text{Pool}(\{\text{GramSchmidt}(\mathbf{X}_{fi})\}_{i=1}^N)$
\For{each atom $i$}
   \State $\mathbf{m}_i^{\text{invariant}} = \mathbf{O}_{\text{global}}^{-1} \mathbf{m}_i$
\EndFor

\State \textbf{Standard diffusion:}
\State Sample $t \sim \text{Uniform}(1, T)$, $\boldsymbol{\epsilon} \sim \mathcal{N}(\mathbf{0}, \mathbf{I})$
\State $\mathbf{z}_t^{\text{invariant}} = \alpha_t \mathbf{m}^{\text{invariant}} + \sigma_t \boldsymbol{\epsilon}$
\State $\hat{\boldsymbol{\epsilon}} = \phi_\theta(\mathbf{z}_t^{\text{invariant}}, t)$
\State Minimize $\mathcal{L} = \|\boldsymbol{\epsilon} - \hat{\boldsymbol{\epsilon}}\|^2$
\end{algorithmic}
\end{algorithm}

When sampling, a sample from the latent noise directly enters the standard diffusion process to generate the molecular output.
No coordinate transformations are required throughout the entire sampling procedure.
This is feasible because the molecule is only globally rotated, and does not need to be transformed back again. 
The output is a valid molecule in Euclidean space, which means it can be directly evaluated. 
In contrast, this approach is not feasible for local frame-based methods, as the predicted outputs of backbone models are in local coordinate systems  and do not correspond to molecule geometries expressed in Euclidean space, making sampling and evaluation impractical.

\begin{algorithm}
\caption{Invariant Frame-based Diffusion (IFD) - Sampling}
\begin{algorithmic}[1]
\Require None
\Ensure Generated canonical molecule $\mathbf{m}^{\text{invariant}}$
 
\State Sample $\mathbf{z}_T \sim \mathcal{N}(\mathbf{0}, \mathbf{I})$
\For{$t = T, T-1, ..., 1$}
    \State $\hat{\boldsymbol{\epsilon}} = \phi_\theta(\mathbf{z}_t, t)$
    \State Sample $\boldsymbol{\epsilon} \sim \mathcal{N}(\mathbf{0}, \mathbf{I})$ if $t > 1$, else $\boldsymbol{\epsilon} = \mathbf{0}$
    \State $\mathbf{z}_{t-1} = \frac{1}{\alpha_{t|t-1}} \mathbf{z}_t - \frac{\sigma_{t|t-1}^2}{\alpha_{t|t-1} \sigma_t} \hat{\boldsymbol{\epsilon}} + \sigma_{t \rightarrow t-1} \boldsymbol{\epsilon}$
\EndFor
\State \Return $\mathbf{z}_0$ (invariant molecule)
\end{algorithmic}
\end{algorithm}

However, the diffusion model cannot generate molecules in different orientations, as all outputs are in a fixed canonical form. 
On the other hand, this allows for both  faster training and generation since we only need to perform the equivariance operation before inputting the diffusion model. 
Since the equivariance operation is performed only once during pre-processing, there is no need for frame transformations during sampling, reducing the computational cost at every time step.
This design aims to resolve the tension between enforcing equivariance and achieving efficiency for practical molecular generation.

For the backbone, we employ Diffusion Transformers (DiT) due to their scalability and expressive capacity. 
To adapt them to molecular geometry, we apply EdgeDiT, a lightweight modification that incorporates interatomic distance and directional cues into the attention mechanism. 
This simple enhancement allows the model to capture chemical bonding patterns more effectively while maintaining the architectural flexibility of Transformer backbones.

\section{Experiments}
\label{sec:experiments}
\subsection{GFD (Main Results)}
We evaluate our proposed frame-based diffusion models on the QM9 dataset, which contains 134k stable small molecules with annotated geometric and chemical properties. Performance is assessed using negative log-likelihood (NLL), atom and molecular stability, validity, and uniqueness.

As shown in Table~\ref{tab:qm9_results}, our GFD model with EdgeDiT achieves decent performance on the QM9 dataset, with test negative log-likelihood of $-137.97$, substantially outperforming all existing methods.
Critically, these improvements are achieved using the same computational scale as the standard SymDiff baseline, indicating that the performance gains stem from methodological innovations rather than increased model capacity. 
This superior test NLL performance indicates that our architecture enables more effective learning, as NLL directly reflects the model's ability to capture the underlying data distribution.

Regarding molecular quality metrics, our model demonstrates superior performance compared to SymDiff at equivalent model scales: atom stability of 98.89\% versus 98.74\%, molecular stability of 89.39\% versus 87.49\%, and validity of 96.04\% versus 95.75\%, while maintaining comparable performance on uniqueness with 97.62\% versus 97.89\%. 
Compared to EDM, the improvements are even more substantial, with molecular stability increasing from 82.00\% to 89.39\% and validity from 91.90\% to 96.04\%. 
Meanwhile, the consistently low variance across all evaluation metrics with standard deviation below $0.03$ further demonstrates the reliability and stability of our deterministic equivariance approach.




\begin{table}[htbp]
\centering
\caption{Test NLL, atom stability, molecular stability, validity and uniqueness on QM9 for 10,000 samples and 3 evaluation runs. * indicates models with double backbone scale. We omit the results for NLL where not available. Bold values indicate the best performance within each model scale/type.}
\label{tab:qm9_results}
\resizebox{\textwidth}{!}{%
\begin{tabular}{lccccc}
\toprule
Method & NLL $\downarrow$ & Atom Stab. (\%) $\uparrow$ & Mol. Stab. (\%) $\uparrow$ & Val. (\%) $\uparrow$ & Uniq. (\%) $\uparrow$ \\
\midrule
\multicolumn{6}{c}{\textit{Other Methods}} \\
\midrule
GeoLDM & -- & $98.90 \pm 0.10$ & $89.40 \pm 0.50$ & $93.80 \pm 0.40$ & $92.70 \pm 0.50$ \\
MUDiff & $\mathbf{-135.50 \pm 2.10}$ & $98.80 \pm 0.20$ & $\mathbf{{89.90 \pm 1.10}}$& $\mathbf{95.30 \pm 1.50}$& ${{99.10 \pm 0.50}}$\\
END & -- & $\mathbf{98.90 \pm 0.00}$ & $89.10 \pm 0.10$ & $94.80 \pm 0.10$ & $92.60 \pm 0.20$ \\
EDM & $-110.70 \pm 1.50$ & $98.70 \pm 0.10$ & $82.00 \pm 0.40$ & $91.90 \pm 0.50$ & $90.70 \pm 0.60$ \\
canon(fr)+GDM & $-104.1$& $97.8$  & $82.1$ & $92.9$ & $99.5$ \\
canon+GDM & ${-117.4 \pm 1.2}$ & ${98.4 \pm 0.2}$ & ${84.6 \pm 0.4}$ & ${94.4 \pm 0.3}$ & $\mathbf{99.7 \pm 0.2}$ \\
\midrule
\multicolumn{6}{c}{\textit{Standard Scale Models}} \\
\midrule
SymDiff & $-129.35 \pm 1.07$ & $98.74 \pm 0.03$ & $87.49 \pm 0.23$ & $95.75 \pm 0.10$ & $97.89 \pm 0.26$ \\
SymDiff-H & $-126.53 \pm 0.90$ & $98.57 \pm 0.07$ & $85.51 \pm 0.18$ & $95.22 \pm 0.18$ & $97.98 \pm 0.09$ \\
DiT-Aug & $-126.81 \pm 1.69$ & $98.64 \pm 0.03$ & $85.85 \pm 0.24$ & $95.10 \pm 0.17$ & $\mathbf{97.98 \pm 0.08}$ \\
DiT & $-127.78 \pm 2.49$ & $98.23 \pm 0.04$ & $81.03 \pm 0.25$ & $94.71 \pm 0.31$ & $97.98 \pm 0.12$ \\
RADM$_{\text{DiT-S}}$ & -- & $98.2 \pm 0.1$ & $83.4 \pm 0.2$ & $92.5 \pm 0.3$ & $90.6 \pm 0.3$  \\
\textbf{GFD (Ours)} & ${\mathbf{-137.97 \pm 0.00}}$ & $\mathbf{98.89 \pm 0.01}$& $\mathbf{89.39 \pm 0.02}$ & $\mathbf{96.04 \pm 0.03}$ & $97.62 \pm 0.01$ \\
\midrule
\multicolumn{6}{c}{\textit{Large Scale Models}} \\
\midrule
SymDiff* (2×) & ${-133.79 \pm 1.33}$& ${{98.92 \pm 0.03}}$& ${89.65 \pm 0.10}$& $\mathbf{{96.36 \pm 0.27}}$& $\mathbf{{97.66 \pm 0.22}}$\\
RADM$_{\text{DiT-B}}$ & -- & ${98.5 \pm 0.0}$ & ${87.3 \pm 0.2}$ & ${94.1 \pm 0.1}$ & ${91.7 \pm 0.1}$ \\
GFD* (Ours) (2×) & $\mathbf{-141.85 \pm 0.00}$& $\mathbf{{98.98 \pm 0.00}}$& $\mathbf{90.51 \pm 0.00}$& ${{96.34 \pm 0.01}}$& ${{97.61 \pm 0.00}}$\\
\midrule
Data & & $99.00$ & $95.20$ & $97.8$ & $100$ \\
\bottomrule
\end{tabular}%
}
\end{table}

To further assess scalability, we double the backbone scale of both GFD and SymDiff. 
As shown in Table~\ref{tab:qm9_results}, GFD* (2×) achieves a test NLL of $-141.85 \pm 0.00$, 
further improving over the standard-scale model by nearly four nats and outperforming SymDiff* by a clear margin. 
Importantly, GFD* also yields the best molecular stability of $90.51\%$, surpassing SymDiff* at $89.65\%$, 
while maintaining strong atom stability ($98.98\%$) and validity ($96.34\%$). 
These results demonstrate that our deterministic equivariance framework not only delivers superior performance 
at standard scale but also scales effectively with increased model capacity, consistently improving generative quality 
without sacrificing stability or efficiency.

We also compare the sampling time of our model with EDM and SymDiff baselines, as shown in Table~\ref{tab:sampling_time}.
While GFD and SymDiff share the same model scale, GFD uses a deterministic equivariant module instead of SymDiff’s stochastic symmetry approach.
Although SymDiff achieves faster sampling speed, GFD still significantly reduces sampling time by about 50\% compared to EDM, while delivering superior generation quality.

\begin{table}[htbp]
\centering
\caption{Sampling time comparison for generating 10,000 molecules.}
\label{tab:sampling_time}
\begin{tabular}{lcc}
\toprule
Method / Seconds per Sample & Mean & Std \\
\midrule
EDM & $0.59694$ & $0.00022$ \\
SymDiff& $0.20731$&$0.00063$\\
canon(fr) + GDF & $0.55$ & -- \\
canon + GDF & $0.55$ & -- \\
GFD (Ours) & $0.32585$ & $0.00119$ \\
\bottomrule
\end{tabular}
\end{table}

\subsection{LFD}
\begin{figure}[htbp]
    \centering
    \includegraphics[width=0.9\linewidth]{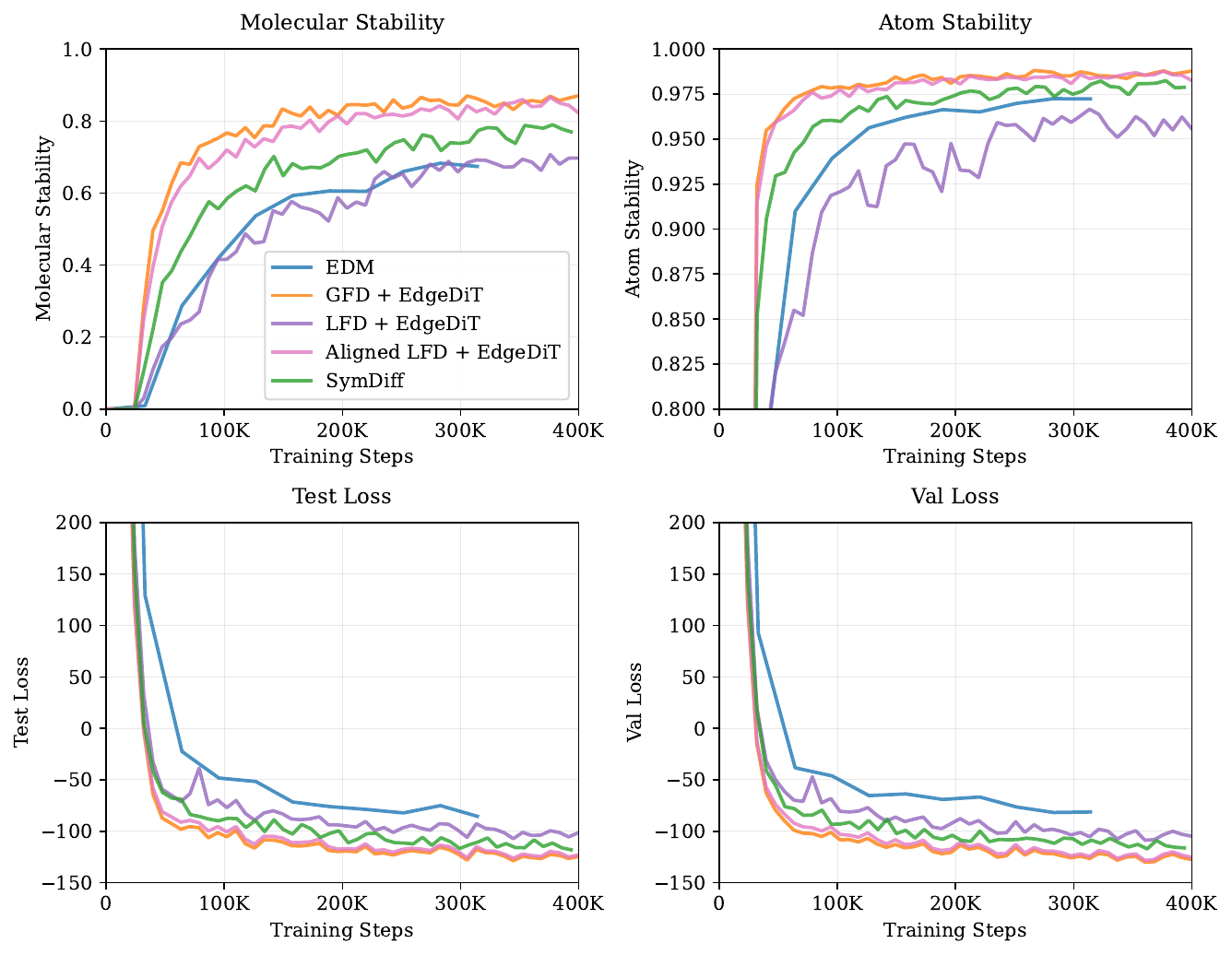}
    \caption{
   Training curves on QM9 comparing EDM, GFD + EdgeDiT, LFD + EdgeDiT, Aligned LFD + EdgeDiT, and SymDiff. Aligned LFD refers to LFD augmented with the proposed frame alignment loss. 
    GFD and Aligned LFD achieve the best overall results, converging faster and attaining higher stability and lower validation loss compared to baselines.
    These results demonstrate that vanilla LFD substantially underperforms GFD, but incorporating frame alignment recovers performance to the level of GFD, thereby validating our hypothesis that preserving global Euclidean structure is essential for effective molecular generation.}
    \label{fig:GFD_LFD_LFD_constrain_comparison}
\end{figure}

On the QM9 dataset, we evaluate our proposed frame-based diffusion models against EDM and SymDiff. 
As shown in Figure~\ref{fig:GFD_LFD_LFD_constrain_comparison}, GFD combined with EdgeDiT achieves consistently superior performance in terms of molecular stability, atom stability, and convergence speed. 
Vanilla LFD, in contrast, significantly underperforms both in stability and likelihood, supporting our hypothesis that preserving the global Euclidean structure is critical for capturing meaningful molecular geometry. 
To further test this hypothesis, we introduced a frame alignment loss that constrains local frames to remain consistent with the global orientation. 
The resulting Aligned LFD model substantially improves over vanilla LFD, matching the performance of GFD in all metrics and convergence behavior. 
These results confirm that retaining global structural consistency is key to effective molecular generation, and that frame alignment serves as a principled mechanism to bridge the gap between local and global representations.

\subsection{IFD}
Building upon the success of the GFD model, we propose another frame-based architecture: Invariant Frame-based Diffusion (IFD).
In this architecture, the canonical step is performed before the diffusion process, which provides faster training and sampling.
Sampling would not involve any canonicalization process and only consider invariant molecules.
We experiment with IFD using EdgeDiT, which has proven to be effective in both IFD and GFD.

As shown in Figure~\ref{fig:frame-invariant}, IFD underperforms GFD in molecular stability, atom stability, and test loss.
While it converges faster initially than EDM, its final performance is worse.
However, the training NLL is better than both GFD and EDM.
So we present it to replace validation loss, which is very similar to test loss.
This might be due to the reduced diversity in invariant molecular representation, which limits the variety of training data for the diffusion model to learn.
This issue does not affect GFD, because GFD learns from every step inside diffusion, allowing it to observe a wider range of noisy invariant molecular representations.

\begin{figure}[htbp]
\centering
\includegraphics[width=\textwidth]{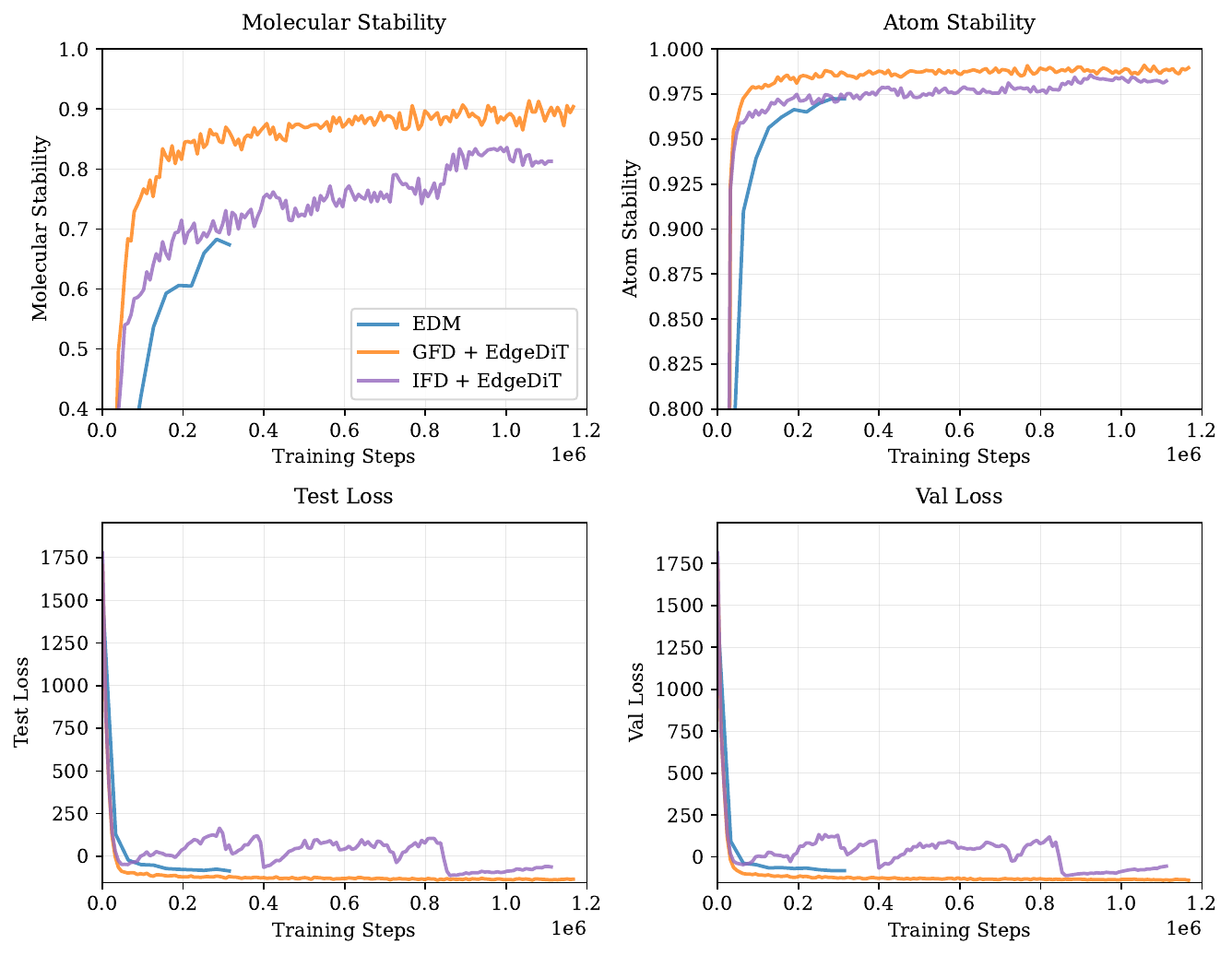}
\caption{Comparison of molecular generation performance across EDM, GFD with EdgeDiT, and IFD with EdgeDiT.}
\label{fig:ifd}
\end{figure}

\section{Discussion and Conclusion}



Our frame-based diffusion paradigm consistently improves molecular generation across evaluation criteria by decoupling deterministic $\mathbb{E}(3)$-equivariance from backbone design. 
On QM9, the Global Frame Diffusion (GFD) model with EdgeDiT achieves state-of-the-art performance, with a test NLL of $-137.97$ at standard scale and $-141.85$ at double scale, while also attaining superior atom stability (98.98\%) and molecular stability (90.51\%) compared to prior equivariant baselines. 
These results confirm that frame-based canonicalization not only guarantees equivariance but also translates into tangible improvements in fidelity, stability, and efficiency, with nearly $2\times$ faster sampling than EDM. 

Our systematic evaluation of three architectural variants provides further insight. 
The Local Frame Diffusion (LFD) variant highlights the importance of structural consistency: without alignment constraints, local frames hinder optimization, but when aligned they markedly improve and approach GFD’s performance, thereby validating our hypothesis. 
These studies emphasize that preserving global structure is essential for effective molecular learning.

The Invariant Frame Diffusion (IFD) variant shows that while strict invariants can ensure equivariance, 
they simultaneously reduce the diversity of geometric views available to the backbone. 
This loss of data diversity limits expressivity and leads to underperformance compared to frame-based canonicalization approaches such as GFD and LFD. 

Despite these strengths, limitations remain. 
GFD introduces canonicalization overhead and relies on MC-EGNN for frame construction, which may constrain scalability on very large datasets or deeper architectures. 
Moreover, our evaluation is currently limited to QM9, and future work should extend to larger and more chemically diverse benchmarks. 
Exploring more efficient frame construction modules or hybrid strategies could further reduce canonicalization costs and enable scaling to foundation-model regimes.

In summary, frame-based diffusion offers a principled and scalable way to enforce deterministic equivariance while leveraging modern high-capacity architectures. 
By systematically comparing GFD, LFD, and IFD, our work establishes the importance of global structure preservation and provides a foundation for future research on scalable, equivariant molecular generative models.




\bibliographystyle{plainnat}
\bibliography{bibentries}

\begin{thebibliography}{25}
\providecommand{\natexlab}[1]{#1}
\providecommand{\url}[1]{\texttt{#1}}
\expandafter\ifx\csname urlstyle\endcsname\relax
  \providecommand{\doi}[1]{doi: #1}\else
  \providecommand{\doi}{doi: \begingroup \urlstyle{rm}\Url}\fi

\bibitem[Alon and Yahav(2021)]{alonbottleneck}
Uri Alon and Eran Yahav.
\newblock On the bottleneck of graph neural networks and its practical implications.
\newblock In \emph{International Conference on Learning Representations}, 2021.
\newblock URL \url{https://openreview.net/forum?id=i80OPhOCVH2}.

\bibitem[Balcilar et~al.(2021)Balcilar, H{\'e}roux, Gauzere, Vasseur, Adam, and Honeine]{balcilar2021breaking}
Muhammet Balcilar, Pierre H{\'e}roux, Benoit Gauzere, Pascal Vasseur, S{\'e}bastien Adam, and Paul Honeine.
\newblock Breaking the limits of message passing graph neural networks.
\newblock In \emph{International Conference on Machine Learning}, pages 599--608. PMLR, 2021.

\bibitem[Blattmann et~al.(2023)Blattmann, Dockhorn, Kulal, Mendelevitch, Kilian, Lorenz, Levi, English, Voleti, Letts, et~al.]{blattmann2023stable}
Andreas Blattmann, Tim Dockhorn, Sumith Kulal, Daniel Mendelevitch, Maciej Kilian, Dominik Lorenz, Yam Levi, Zion English, Vikram Voleti, Adam Letts, et~al.
\newblock Stable video diffusion: Scaling latent video diffusion models to large datasets.
\newblock \emph{CoRR}, 2023.

\bibitem[DiMasi et~al.(2016)DiMasi, Grabowski, and Hansen]{dimasi2016innovation}
Joseph~A DiMasi, Henry~G Grabowski, and Ronald~W Hansen.
\newblock Innovation in the pharmaceutical industry: new estimates of r\&d costs.
\newblock \emph{Journal of health economics}, 47:\penalty0 20--33, 2016.

\bibitem[Ding and Hofmann(2025)]{dingscalable}
Yuhui Ding and Thomas Hofmann.
\newblock Scalable non-equivariant 3d molecule generation via rotational alignment.
\newblock In \emph{Forty-second International Conference on Machine Learning}, 2025.

\bibitem[Fang et~al.(2022)Fang, Liu, Lei, He, Zhang, Zhou, Wang, Wu, and Wang]{fang2022geometry}
Xiaomin Fang, Lihang Liu, Jieqiong Lei, Donglong He, Shanzhuo Zhang, Jingbo Zhou, Fan Wang, Hua Wu, and Haifeng Wang.
\newblock Geometry-enhanced molecular representation learning for property prediction.
\newblock \emph{Nature Machine Intelligence}, 4\penalty0 (2):\penalty0 127--134, 2022.

\bibitem[Ho et~al.(2020)Ho, Jain, and Abbeel]{ho2020denoising}
Jonathan Ho, Ajay Jain, and Pieter Abbeel.
\newblock Denoising diffusion probabilistic models.
\newblock \emph{Advances in Neural Information Processing Systems}, 33:\penalty0 6840--6851, 2020.

\bibitem[Ho et~al.(2022)Ho, Salimans, Gritsenko, Chan, Norouzi, and Fleet]{ho2022video}
Jonathan Ho, Tim Salimans, Alexey Gritsenko, William Chan, Mohammad Norouzi, and David~J Fleet.
\newblock Video diffusion models.
\newblock \emph{Advances in Neural Information Processing Systems}, 35:\penalty0 8633--8646, 2022.

\bibitem[Hoogeboom et~al.(2022)Hoogeboom, Satorras, Vignac, and Welling]{hoogeboom2022equivariant}
Emiel Hoogeboom, V{\i}ctor~Garcia Satorras, Cl{\'e}ment Vignac, and Max Welling.
\newblock Equivariant diffusion for molecule generation in 3d.
\newblock In \emph{International Conference on Machine Learning}, pages 8867--8887. PMLR, 2022.

\bibitem[Kaba et~al.(2023)Kaba, Mondal, Zhang, Bengio, and Ravanbakhsh]{kaba2023equivariance}
S{\'e}kou-Oumar Kaba, Arnab~Kumar Mondal, Yan Zhang, Yoshua Bengio, and Siamak Ravanbakhsh.
\newblock Equivariance with learned canonicalization functions.
\newblock In \emph{International Conference on Machine Learning}, pages 15546--15566. PMLR, 2023.

\bibitem[Levy et~al.(2023)Levy, Kaba, Gonzales, Miret, and Ravanbakhsh]{levy-icml2023}
Daniel Levy, Sékou-Oumar Kaba, Carmelo Gonzales, Santiago Miret, and Siamak Ravanbakhsh.
\newblock Using multiple vector channels improves e (n)-equivariant graph neural networks.
\newblock In \emph{ICML Workshop on Machine Learning for Astrophysics}, 2023.
\newblock URL \url{https://ml4astro.github.io/icml2023/assets/68.pdf}.

\bibitem[Lippmann et~al.(2025)Lippmann, Gerhartz, Remme, and Hamprecht]{lippmann2025beyond}
Peter Lippmann, Gerrit Gerhartz, Roman Remme, and Fred~A Hamprecht.
\newblock Beyond canonicalization: How tensorial messages improve equivariant message passing.
\newblock In \emph{The Thirteenth International Conference on Learning Representations}, 2025.

\bibitem[Luo et~al.(2021)Luo, Shi, Xu, and Tang]{luo2021predicting}
Shitong Luo, Chence Shi, Minkai Xu, and Jian Tang.
\newblock Predicting molecular conformation via dynamic graph score matching.
\newblock \emph{Advances in Neural Information Processing Systems}, 34:\penalty0 19784--19795, 2021.

\bibitem[Luo et~al.(2022)Luo, Li, Guan, Su, Cheng, Peng, and Ma]{Luo_2022_CVPR}
Shitong Luo, Jiahan Li, Jiaqi Guan, Yufeng Su, Chaoran Cheng, Jian Peng, and Jianzhu Ma.
\newblock Equivariant point cloud analysis via learning orientations for message passing.
\newblock In \emph{Proceedings of the IEEE/CVF Conference on Computer Vision and Pattern Recognition (CVPR)}, pages 18932--18941, June 2022.

\bibitem[Ma et~al.(2024)Ma, Wang, Lim, Jegelka, and Wang]{ma2024a}
George Ma, Yifei Wang, Derek Lim, Stefanie Jegelka, and Yisen Wang.
\newblock A canonicalization perspective on invariant and equivariant learning.
\newblock In \emph{The Thirty-eighth Annual Conference on Neural Information Processing Systems}, 2024.
\newblock URL \url{https://openreview.net/forum?id=jjcY92FX4R}.

\bibitem[Peebles and Xie(2023)]{peebles2023scalable}
William Peebles and Saining Xie.
\newblock Scalable diffusion models with transformers.
\newblock In \emph{Proceedings of the IEEE/CVF international conference on computer vision}, pages 4195--4205, 2023.

\bibitem[Rombach et~al.(2022)Rombach, Blattmann, Lorenz, Esser, and Ommer]{rombach2022high}
Robin Rombach, Andreas Blattmann, Dominik Lorenz, Patrick Esser, and Bj{\"o}rn Ommer.
\newblock High-resolution image synthesis with latent diffusion models.
\newblock In \emph{Proceedings of the IEEE/CVF conference on computer vision and pattern recognition}, pages 10684--10695, 2022.

\bibitem[Sareen et~al.(2025)Sareen, Levy, Mondal, Kaba, Akhound-Sadegh, and Ravanbakhsh]{DBLP:journals/corr/abs-2501-07773}
Kusha Sareen, Daniel Levy, Arnab~Kumar Mondal, Sékou-Oumar Kaba, Tara Akhound-Sadegh, and Siamak Ravanbakhsh.
\newblock Symmetry-aware generative modeling through learned canonicalization.
\newblock \emph{CoRR}, abs/2501.07773, January 2025.
\newblock URL \url{https://doi.org/10.48550/arXiv.2501.07773}.

\bibitem[Satorras et~al.(2021)Satorras, Hoogeboom, and Welling]{satorras2021n}
V{\'i}ctor~Garcia Satorras, Emiel Hoogeboom, and Max Welling.
\newblock E (n) equivariant graph neural networks.
\newblock In \emph{International Conference on Machine Learning}, pages 9323--9332. PMLR, 2021.

\bibitem[Scarselli et~al.(2008)Scarselli, Gori, Tsoi, Hagenbuchner, and Monfardini]{scarselli2008graph}
Franco Scarselli, Marco Gori, Ah~Chung Tsoi, Markus Hagenbuchner, and Gabriele Monfardini.
\newblock The graph neural network model.
\newblock \emph{IEEE transactions on neural networks}, 20\penalty0 (1):\penalty0 61--80, 2008.

\bibitem[Song and Ermon(2019)]{song2019generative}
Yang Song and Stefano Ermon.
\newblock Generative modeling by estimating gradients of the data distribution.
\newblock \emph{Advances in Neural Information Processing Systems}, 32, 2019.

\bibitem[Xu et~al.(2022)Xu, Yu, Song, Shi, Ermon, and Tang]{xu2022geodiff}
Minkai Xu, Lantao Yu, Yang Song, Chence Shi, Stefano Ermon, and Jian Tang.
\newblock Geodiff: A geometric diffusion model for molecular conformation generation.
\newblock In \emph{International Conference on Learning Representations}, 2022.
\newblock URL \url{https://openreview.net/forum?id=PzcvxEMzvQC}.

\bibitem[Xu et~al.(2023)Xu, Powers, Dror, Ermon, and Leskovec]{xu2023geometric}
Minkai Xu, Alexander Powers, Ron Dror, Stefano Ermon, and Jure Leskovec.
\newblock Geometric latent diffusion models for 3d molecule generation.
\newblock In \emph{International Conference on Machine Learning}. PMLR, 2023.

\bibitem[Zhang et~al.(2025)Zhang, Ashouritaklimi, Teh, and Cornish]{zhang2025symdiff}
Leo Zhang, Kianoosh Ashouritaklimi, Yee~Whye Teh, and Rob Cornish.
\newblock Symdiff: Equivariant diffusion via stochastic symmetrisation.
\newblock In \emph{The Thirteenth International Conference on Learning Representations}, 2025.
\newblock URL \url{https://openreview.net/forum?id=i1NNCrRxdM}.

\bibitem[Zhou et~al.(2023)Zhou, Gao, Ding, Zheng, Xu, Wei, Zhang, and Ke]{zhou2023uni}
Gengmo Zhou, Zhifeng Gao, Qiankun Ding, Hang Zheng, Hongteng Xu, Zhewei Wei, Linfeng Zhang, and Guolin Ke.
\newblock Uni-mol: A universal 3d molecular representation learning framework.
\newblock In \emph{The Eleventh International Conference on Learning Representations}, 2023.
\newblock URL \url{https://openreview.net/forum?id=6K2RM6wVqKu}.

\end{thebibliography}







\appendix

\section{Implementation details}

Our models are configured to align closely with baseline settings for fair comparison.
Specifically, MC EGNN consists of  3 layers, 7 internal channels, and a hidden size of 256.
MPNN uses 9 layers, and 256-dimensional hidden features.
Transformer and all its variants have 9 layers, 8 heads, and hidden size of 256.
DiT and all its variants have 12 layers, 6 heads, and hidden size of 384.
Experiments with MPNN and Transformer backbones are trained with learning rate 0.0001 and batch size 64.
Experiments with DiT backbones are trained with learning rate 0.0002 and batch size 256.
All experiments are conducted with AdamW optimizer on NVIDIA A100 GPUs with 40GB memory.


\end{document}